%% file: iclr2027_conference.tex
\documentclass{article} 
\usepackage{iclr2027_conference,times}

\input{math_commands.tex}

\usepackage{hyperref}
\hypersetup{pdfborder={0 0 0}}
\usepackage{url}
\usepackage[utf8]{inputenc}
\usepackage[T1]{fontenc}
\usepackage{booktabs}
\usepackage{amsfonts}
\usepackage{nicefrac}
\usepackage{microtype}
\usepackage{xcolor}
\usepackage{graphicx}
\usepackage{amsmath}
\usepackage{amssymb}
\usepackage{booktabs}
\usepackage{multirow}
\usepackage{graphicx}
\usepackage{caption}
\usepackage{enumitem}
\usepackage{adjustbox}

\title{FlashLoop: Fast and Memory-Efficient Looped \\
Transformers via Lazy Updates}

\author{Wanqi Yang \\
ELLIS Institute T\"ubingen  \\
Max Planck Institute for Intelligent Systems \\
T\"ubingen AI Center \\
\And
Shiwei Liu \\
ELLIS Institute T\"ubingen  \\
Max Planck Institute for Intelligent Systems \\
T\"ubingen AI Center
}

\iclrfinalcopy
\begin{document}

\maketitle

\begin{abstract}
Looped Transformers have attracted substantial attention as a parameter-efficient approach to increasing computational depth through repeated application of shared Transformer blocks. However, their practical advantages over conventional Transformers remain under debate: each additional loop incurs another Transformer pass and requires caching another set of KV states, causing inference FLOPs and KV-cache memory to grow continuously with loop depth. This overhead becomes particularly severe at large loop counts and long context, preventing the parameter efficiency of Looped Transformers from translating into practical inference efficiency.
In this paper, we find that much of the additional computation and storage introduced by looping is redundant. As recurrence proceeds, state changes become increasingly concentrated on a small subset of tokens; attention-output differences are dominated by a sparse and stable subset of key columns; and KV residuals between adjacent loops become progressively more amenable to low-bit quantization. Building on these observations, we introduce \textbf{FlashLoop}, a training-free inference framework that reduces cross-loop redundancy through token-sparse updates, sparse attention, and KV-residual quantization. Across several Looped Transformers models, \textsc{FlashLoop} delivers lossless accuracy while achieving up to \textbf{1.64}$\times$ end-to-end speedup and up to \textbf{6}$\times$ KV-cache memory reduction, substantially improving the practicality of scaling Looped Transformers to greater computational depths and longer context.

\par
{\small
\noindent\textbf{Project Page:}
\href{https://superone77.github.io/FlashLoop/}{\textcolor{blue!65!black}{\texttt{https://superone77.github.io/FlashLoop/}}}\\
\textbf{Code:}
\href{https://github.com/Superone77/FlashLoop}{\textcolor{blue!65!black}{\texttt{https://github.com/Superone77/FlashLoop}}}
}

\end{abstract}

\section{Introduction}
Looped Transformers offer a promising way to decouple a model's computational depth from its parameter count. By repeatedly applying a shared Transformer module, they allow relatively compact models to perform substantially deeper computation while keeping the stored model weights fixed~\citep{dehghani2018universal,zhu2025scaling,geiping2026scaling}. This opens a new scaling dimension: increasing test-time compute and thereby computational depth, without increasing the model's parameter-memory footprint.

However, a lower parameter count does not necessarily translate into lower inference costs. Unrolling an \(N\)-layer looped Transformer with \(R\) loops requires \(R \times N\) block executions and a distinct KV cache per iteration, making computation and KV-cache memory grow linearly with loop depth. On a single A100 machine, the prefill time for a 32K-token prompt is 27 seconds for Ouro-2.6B \citep{zhu2025scaling} with four loops, compared with only 3 seconds for LLaMA-3.1-8B \citep{grattafiori2024llama}. Meanwhile, Ouro's KV cache alone requires 48 GiB, compared with just 4 GiB for LLaMA-3.1-8B. Thus, despite having roughly one-third as many parameters, the looped model can impose substantially greater latency and memory requirements.

To uncover opportunities for reducing this growing inference overhead, we systematically characterize the cross-loop dynamics of representative looped Transformers. We observe a pronounced \emph{lazy-update pattern} in their inference dynamics: as recurrence progresses, changes between adjacent loops become increasingly sparse, predictable, and compressible. This pattern gives rise to three major forms of redundancy:
\begin{itemize}[
    leftmargin=1.2em,
    labelsep=0.4em,
    itemsep=1pt,
    topsep=2pt,
    parsep=0pt,
    partopsep=0pt
]
    \item Token-update redundancy. Updates to most token states become increasingly small, with only a small fraction of token rows accounting for the majority of changes in hidden states and KV representations.
    \item Attention-computation redundancy. Cross-loop differences in attention outputs are dominated by a sparse subset of columns whose indices can be reliably predicted from the preceding loop.
    \item KV-storage redundancy. KV residuals between adjacent loops are substantially more friendly to quantization than full KV states, exhibiting lower reconstruction error under identical quantization settings.
\end{itemize}

\begin{figure}
    \centering
    \includegraphics[width=\linewidth]{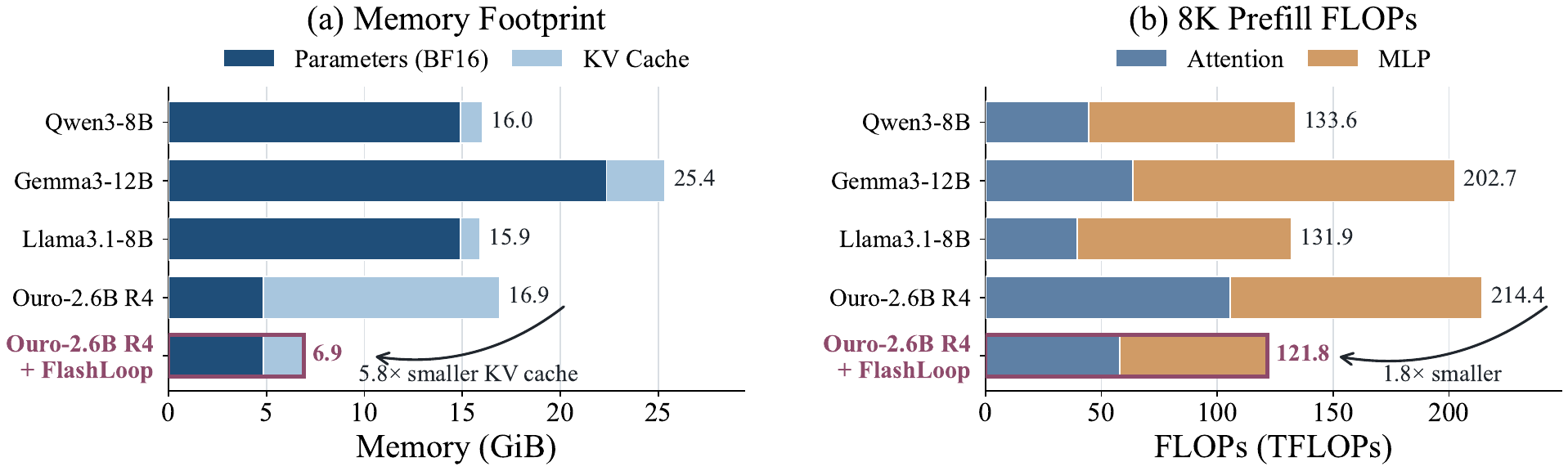}
    \caption{\textbf{Comparison of performance-matched looped and non-looped models at an 8K context length\protect\footnotemark.}
    FlashLoop substantially reduces both computation and KV-cache memory for looped Transformers, achieving 5.8$\times$ and 1.8$\times$ reductions in KV cache memory and 8K prefill FLOPs.}
    \label{fig:pareto_frontier}
\end{figure}
\footnotetext{Accuracy results for these models are reported in Appendix~\ref{app:performance_open_sourced}.}

Building on these observations, we introduce \textbf{FlashLoop}, a training-free and loop-native inference framework that reduces cross-loop redundancy through three  complementary components. First, \emph{Token-Sparse Updates} reuses the hidden and KV states of converged tokens, recomputing states and storing new KV entries only for active tokens. This reduces computation and avoids maintaining a complete KV cache for every loop. Second, \emph{Sparse Attention} selects important keys using attention-contribution statistics from the preceding loop, while cached global probability mass preserves the full-softmax scale. Third, \emph{KV-Residual Quantization} compresses the cache into a quantized base state and low-bit residual updates between successive loops. We implement these components efficiently on GPUs through hardware-aware pipelining and optimized kernels.

\looseness=-1 Across several Looped Transformers models, FlashLoop retains the parameter efficiency of looped Transformers and largely preserves their accuracy, while delivering up to $1.64\times$ end-to-end speedup and reducing KV-cache memory by up to $6\times$. As illustrated in Figure~\ref{fig:pareto_frontier}, these improvements bring the inference costs of looped transformers to levels comparable to, or lower than, those of performance-matched non-looped models.

\section{Related Work}
\label{sec:related-work}

\paragraph{Looped Transformers and loop-specific KV-cache reduction.}
The Universal Transformer introduced depth recurrence and adaptive per-position computation to combine the parallelism of Transformers with the inductive biases of recurrent models~\citep{dehghani2018universal}.  Subsequent recurrent language models develop this idea in different ways. Chain-of-Experts sequentially reuses FFN experts with iteration-specific routing to increase effective computational depth~\citep{wang2025chain}. Huginn repeatedly applies a shared recurrent module and treats the recurrence count as a potential control over test-time compute~\citep{geiping2026scaling}. Ouro instead incorporates latent reasoning capacity into pretraining through weight-shared recurrence, showing high parameter efficiency~\citep{zhu2025scaling}.
Several methods have been proposed to manage the resulting loop-indexed KV cache. Memory-Efficient Looped Transformer reduces cache overhead through learned gated updates, whereas Looped Latent Attention trains encoders and decoders to construct compressed representations along the recurrence dimension~\citep{vendrell2026memory,neill2026looped}. However, these methods either modify the model architecture or require additional training.
In contrast, FlashLoop starts from a pretrained recurrent model, eliminates cross-loop structural redundancy at inference time without modifying model weights or requiring any training.

\paragraph{Generic inference optimization.}
Existing general inference optimization methods reduce computational and memory overhead along several complementary dimensions. Quantization lowers the storage and bandwidth costs of the parameters and KV cache~\citep{liu2024kivi,hooper2024kvquant, yang2026alphaq}. KV cache sharing merges or reuses keys and values by exploiting similarities or redundancies across network layers~\citep{liu2024minicache,brandon2024reducing}. Token and KV eviction controls the effective context length and cache overhead by retaining only the historical tokens deemed most important for subsequent generation~\citep{liu2023scissorhands,zhang2023h2o,li2024snapkv,cai2024pyramidkv}. Sparse attention reduces attention computation and KV reads through structured attention patterns or query-dependent selection~\citep{ribar2023sparq,jiang2024minference,tang2024quest}.
Collectively, these methods primarily target non-looped Transformers and therefore do not explicitly exploit structural redundancy along the recurrence axis of looped Transformers. In contrast, FlashLoop identifies loop-specific, multilevel redundancy arising during looped execution and provides targeted optimizations for this redundancy, making it complementary to existing general inference optimization approaches.
\begin{figure}
    \centering
    \captionsetup{skip=5pt}    \includegraphics[width=1\linewidth]{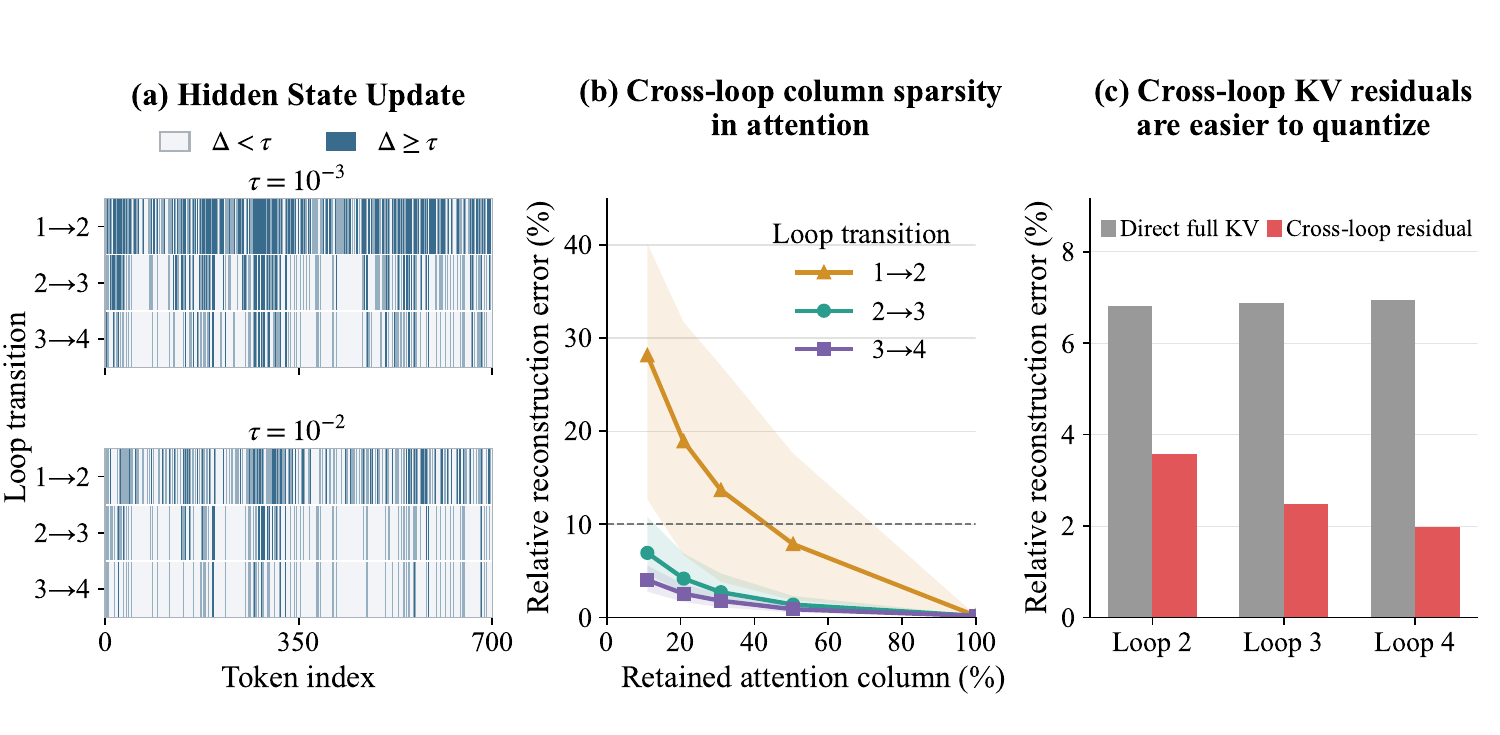}
    \caption{\textbf{Recurrent refinement becomes structurally redundant along three complementary axes.}
  (a) \emph{Token-level cross-loop redundancy}: hidden-state change becomes increasingly concentrated at a subset of tokens.
  (b) \emph{Attention-column redundancy}: later-loop attention outputs can be reconstructed from a small subset of key columns identified by the preceding loop.
  (c) \emph{Cross-loop representation redundancy}: at the same quantization setting, recursively quantizing adjacent-loop KV residuals gives lower reconstruction error than independently quantizing complete KV states.}
  \label{fig:flashloop_motivation}
\end{figure}
\section{Cross-Loop Redundancy in Looped Transformers}
\label{sec:cross-loop-redundancy}
Let one recurrent computation of the looped Transformer parameters be defined as
\begin{equation}
H^{(r)} = F_{\theta}\left(H^{(r-1)}\right),
\end{equation}
where $r\in{1,\ldots,R}$ denotes the loop index, $H^{(r)}$ denotes the token hidden states after the $r$-th loop, and $F_{\theta}$ denotes the Transformer block parameterized by $\theta$ and shared across different loops. In the original implementation, $F_{\theta}$ is applied to all tokens at every loop, while a separate pair of keys and values, $\left(K^{(r)},V^{(r)}\right)$, is stored for each loop, which means the computation and KV cache increase linearly with loop count.

This execution pattern, however, treats every loop as a full update, overlooking the increasingly structured changes between successive loops. Specifically, substantial hidden-state changes become concentrated in a small subset of token rows, important attention columns remain concentrated and stable across loops, and adjacent-loop KV residuals are more amenable to quantization than full KV states. Figure~\ref{fig:flashloop_motivation} and Appendix \ref{app:redundancy_in_more_loops} summarize these three forms of cross-loop structural redundancy.

\subsection{Token-Update Redundancy}
\label{sec:token_redundancy}
The first form of redundancy we observe is that iterative refinement is non-uniform across token positions in the context: the hidden states of different tokens converge at different rates across loops. Figure~\ref{fig:flashloop_motivation}(a) illustrates two key characteristics of token-level convergence across loops. First, tokens with substantial hidden-state changes become increasingly sparse as recurrence proceeds, while remaining distributed throughout the context. Second, these high-change tokens exhibit an approximately nested structure: tokens that remain active in a later loop transition are largely a subset of those active in the preceding transition. 

Since keys and values are computed independently from each token representation, this token-level sparsity of hidden-state updates naturally induces sparsity in KV updates. Thus, tokens whose hidden states have effectively converged can directly reuse their KV states from the preceding loop. This progressive concentration of token-level state changes motivates the sparse computation mechanism in FlashLoop, which selectively shares KV cache across loops and updates tokens based on their cross-loop changes.

\begin{figure}
    \centering
    \includegraphics[width=1\linewidth]{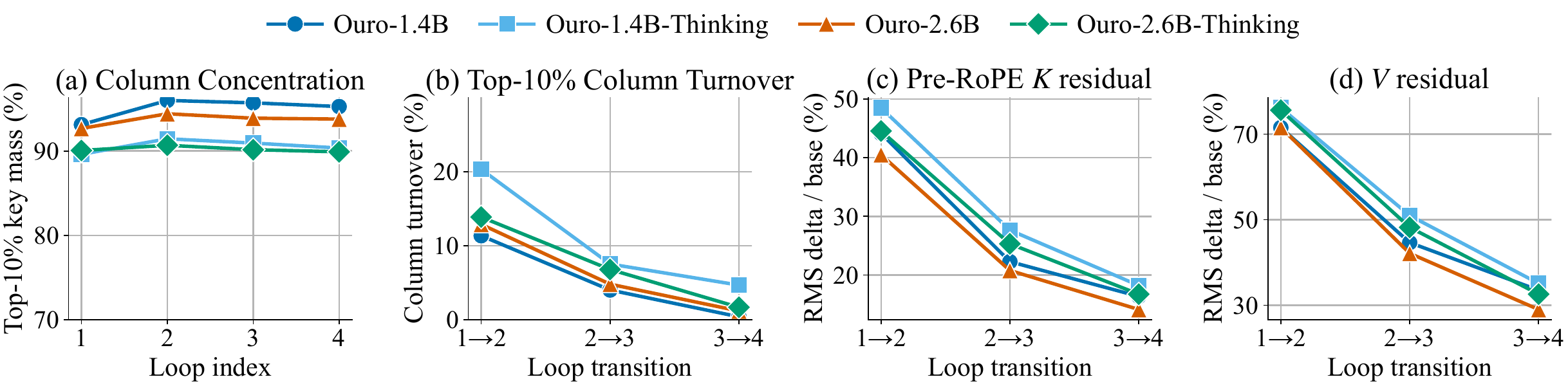}
    \caption{\textbf{Additional evidence for cross-loop redundancy across different looped models.} (a) Different looped transformers have similar cross-loop redundancy: decode attention concentrates on a small set of attention columns, (b) whose importance becomes increasingly stable across loops, (c--d) while adjacent-loop K/V residuals decrease with loop depth motivating a residual-based quantization for cross-loop KV states.   }
    \label{fig:cross_loop_redundancy}
\end{figure}

\subsection{Attention-Computation Redundancy}
\label{sec:column_redundancy}
The second form of redundancy we observe lies in the cross-loop updates of the attention output. Let’s revisit the equation of attention:
\begin{equation}
    o^{(r)}
    =
    \operatorname{softmax}
    \left(
        \frac{
            q^{(r)}
            {K^{(r)}}^{\top}
        }{
            \sqrt{d}
        }
    \right)
    V^{(r)} .
\label{eq:attention-output}
\end{equation}
Each key/value index corresponds to a column of the attention matrix across queries, whose attention weights scale the associated value vector in the output aggregation. We refer to this key/value-specific contribution as an \emph{attention column}. We find that cross-loop changes in the attention output are concentrated on a sparse subset of attention columns: most change only slightly, while a small subset accounts for most of the change in the attention output, as shown in Figure~\ref{fig:cross_loop_redundancy}(a). Moreover, the importance of these columns remains stable across adjacent loops (Figure~\ref{fig:cross_loop_redundancy}(b), Figure \ref{fig:appendix_attention_maps}), allowing the previous loop to identify the columns that are likely to remain important in the next loop.

This observation suggests approximating the current-loop attention output with sparse column-wise updates to the previous-loop output. Leveraging the cross-loop stability of column importance, we rank key columns by their attention probabilities in loop $r-1$ and let $S$ denote the indices of the top-$K$ columns. We then reconstruct the attention output of loop $r$ by replacing only these columns' contributions in $o^{(r-1)}$:
\begin{equation}
\resizebox{\dimexpr\linewidth-3em\relax}{!}{$\displaystyle
    \widehat{o}^{(r)}
    = o^{(r-1)}
    -
    m\left[
        \operatorname{softmax}
        \left(
            \frac{q^{(r-1)}{K_S^{(r-1)}}^\top}{\sqrt{d}}
        \right)
    \right]
    V_S^{(r-1)}
    +
    m\left[
        \operatorname{softmax}
        \left(
            \frac{q^{(r)}{K_S^{(r)}}^\top}{\sqrt{d}}
        \right)
    \right]
    V_S^{(r)}
$}
\label{eq:sparse-attention-delta}
\end{equation}
where $m$ is a renormalization factor used to preserve the global attention mass of the selected columns.
The selected columns are recomputed using the current-loop states, while the remaining contributions are reused from the previous loop.
Figure~\ref{fig:flashloop_motivation}(b) shows that the reconstruction error decreases as more columns are retained and, for the same retention rate, becomes substantially lower at later loop transitions. In loop 3/4 of Ouro-1.4B, updating only 10\% of the columns can reconstruct over 90\% of exact attention output.

\subsection{Quantization-Friendly KV Residual}
\label{sec:residual_friendly}
Although our previous observations allow a subset of tokens to be set as inactive in the middle and later loops and reuse their previously generated KV cache, active tokens continue to produce new KV cache in both early and later loops. In long-context settings, caching these KV states accounts for a major portion of the memory overhead. 

\looseness=-1 To compress this part of the KV cache through quantization, we analyze its quantization friendliness of the full KV states and the KV residual between the current loop and their preceding loop. As shown in Figure~\ref{fig:cross_loop_redundancy}(c,d), the magnitudes of these residuals decrease with loop depth, making the incremental updates increasingly small relative to the full KV states and thus more sensitive to quantization error when the full states are quantized directly. The results in Figure~\ref{fig:flashloop_motivation}(c) show that quantizing these cross-loop residuals yields substantially lower relative reconstruction error than directly quantizing the full KV states, and this advantage becomes more pronounced as the loop depth increases. These observations motivate representing the loop-wise KV cache as a quantized base state with quantized cross-loop residual updates.

\section{FlashLoop}
\label{sec:method}

To reduce these structural redundancies, we propose FlashLoop, a training-free and loop-native inference framework for pretrained looped Transformers. FlashLoop includes three coupled compression strategies: token-sparse updates, sparse attention, and KV residual quantization. We also provide the mathematical formulation of our framework in Appendix \ref{app:flashloop_math}.

\subsection{Cross-Loop Token-Sparse Updates}
\label{sec:cross_loop_kv_sharing}

Motivated by the progressive concentration of hidden-state changes observed in Section~\ref{sec:token_redundancy}, FlashLoop selectively updates only the tokens that remain active across loops. After the dense early loops, each subsequent loop ranks the tokens updated in the preceding loop according to their normalized hidden-state changes and retains only the top-$k$ tokens for further refinement, where $k$ can vary with loop depth. Since skipped tokens do not produce new hidden-state changes, later selections are restricted to the active tokens from the preceding loop, naturally yielding nested active-token sets.

This design reduces both computation and memory overhead. For each sparse loop, the looped Transformer block performs computation only on the selected active tokens, which avoids wasting computation on those inactive tokens. The same token-level decision also applies to KV-cache updates. Active tokens generate new K/V states for the current loop, whereas inactive tokens continue to reuse their KV cache from preceding loop. Consequently, a later-loop KV cache does not require another complete copy of all token states and saves memory for that.

\subsection{Loop-Aware Sparse Attention}
\label{sec:sparse-attention}

Building on the cross-loop sparsity and stability of important attention columns identified in Section~\ref{sec:column_redundancy}, FlashLoop uses the preceding loop to identify a small subset of columns for recomputation. For each sparse loop, we rank the attention columns using statistics from the previous loop and select the top-\(K\) columns under a loop-specific budget. We then gather only their corresponding key/value entries, allowing both the QK product and value aggregation to operate on this reduced set.

A direct softmax over the selected columns, however, would renormalize their probabilities to sum to one and therefore overestimate their contribution to the full attention output. Thus, FlashLoop caches their \emph{global probability mass}, i.e., the total probability assigned to the selected columns under the full attention distribution of the preceding loop. In the target loop, we apply softmax only over the selected logits and rescale the resulting probabilities by this cached mass, approximately preserving their contribution under the full attention distribution.

The rescaled value contributions are then used to update the attention output according to Eq.~\ref{eq:sparse-attention-delta}. When later loops use smaller column budgets, we further prune the selected set using attention statistics from the preceding sparse loop. Both the selected column indices and their associated probability mass are propagated across loops, enabling progressively sparser attention computation.

\subsection{Cross-Loop KV Residual Quantization}
\label{sec:kv-quantization}

As observed in Section~\ref{sec:residual_friendly}, cross-loop KV residuals are more quantization-friendly than full KV states. FlashLoop therefore represents the KV cache as a quantized base state followed by quantized residual updates across loops. The K/V states from the first loop are quantized as the base, while each subsequent loop stores only the residual relative to the reconstructed state of the preceding loop. For $X\in\{K,V\}$, we define
\begin{equation}
\begin{aligned}
    \widehat{X}^{(1)}
    &= D\!\left(Q\!\left(X^{(1)}\right)\right),
    \qquad
    \Delta X^{(r)}
    = X^{(r)}-\widehat{X}^{(r-1)}, \\
    \widehat{X}^{(r)}
    &= \widehat{X}^{(r-1)}
    + D\!\left(Q\!\left(\Delta X^{(r)}\right)\right),
    \qquad 2\leq r\leq R ,
\end{aligned}
\label{eq:kv-residual-quantization}
\end{equation}
where $Q(\cdot)$ and $D(\cdot)$ denote quantization and dequantization, respectively. At read time, the required K/V state is reconstructed by accumulating the quantized residual updates on top of the base state.

Cross-loop KV residual quantization naturally complements our cross-loop token-sparse updates. In sparse loops, inactive tokens reuse their existing K/V states, while residuals are generated and stored only for active tokens. FlashLoop therefore reduces both the number of new K/V entries stored across loops and the storage and bandwidth cost of each remaining update. 

\subsection{Fast GPU Support}
\label{sec:gpu-support}
\looseness=-1 Although the blueprint described above is simple at the algorithmic level, efficiently realizing sparsity and low-bit KV residual on GPUs requires careful system support. Dynamic token and column selection can introduce irregular memory accesses and fragmented computation, while separate cache-update and quantization steps may add additional data movement and kernel-launch overhead. FlashLoop addresses these issues by gathering selected tokens and keys into compact, compute-friendly buffers to preserve regular GEMM execution, and by fusing cross-loop KV updates and quantization into the attention kernel. These optimizations allow FlashLoop to translate its algorithmic sparsity and compression into practical reductions in computation and KV-cache memory. We provide more details in Appendix \ref{app:open_source}.

\section{Experiments}
\subsection{Experimental Setup}

\label{sec:experimental-setup}

We evaluate FlashLoop on five representative Looped Transformers: four models from the Ouro \citep{zhu2025scaling} family and Huginn-3.5B \citep{geiping2026scaling}. We evaluate FlashLoop on five benchmarks: MATH-500~\citep{lightman2023lets}, GSM8K~\citep{cobbe2021gsm8k}, ARC-Challenge~\citep{allenai_arc}, HellaSwag~\citep{zellers2019hellaswag} and WinoGrande~\citep{sakaguchi2019winogrande}, using the EleutherAI LM Harness~\citep{eval-harness}. Moreover, we provide the detailed setting such as token and column sparsity settings for each loop of each model and the quantization setting in Appendix~\ref{app:experimental_setting}.

\subsection{Main Results}
\label{sec:main-results}

We report our results in Table~\ref{tab:main-quality}. Overall, FlashLoop
largely preserves the task performance of the original models while
substantially reducing their inference costs. Moreover, across model variants, base models tend to be more robust to FlashLoop
than their thinking counterparts, suggesting that thinking models may be
more sensitive to cross-loop compression. Nevertheless, FlashLoop still
largely preserves the performance of the thinking models, indicating that
substantial cross-loop redundancy can be exploited even in these models.

\begin{table*}[t]
\centering
\small
\setlength{\tabcolsep}{4.5pt}
\caption{\textbf{Task performance and efficiency of Ouro and Huginn with FlashLoop.}
We report accuracy (\%) on five benchmarks.
$\Delta$ Avg. denotes the average-score difference from the original model
in percentage points. KV Memory means the reduction of peak KV cache memory relative to original models.}
\label{tab:main-quality}
\resizebox{\textwidth}{!}{%
\begin{tabular}{@{}llccccccccc@{}}
    \toprule
    Model & Method & MATH-500 & ARC-C & GSM8K & HellaSwag
          & WinoGrande & Avg. & $\Delta$ Avg. & Speedup $\uparrow$ & KV Memory $\uparrow$ \\
    \midrule

    \multirow{2}{*}{Ouro-1.4B}
        & Original
        & 65.80 & 59.98 & 78.77
        & 74.24 & 71.74 & 70.11 & --
        & $1.00\times$ & $1.00\times$ \\
        & FlashLoop
        & 65.40 & 60.67 & 79.61
        & 74.33 & 72.38 & 70.48 & $+0.37$
        & $1.59\times$ & $5.85\times$ \\

    \midrule
    \multirow{2}{*}{Ouro-1.4B-Thinking}
        & Original
        & 46.80 & 62.03 & 81.65
        & 72.61 & 72.45 & 67.11 & --
        & $1.00\times$ & $1.00\times$ \\
        & FlashLoop
        & 46.20 & 61.69 & 80.39
        & 72.38 & 70.51 & 66.23 & $-0.88$
        & $1.59\times$ & $5.85\times$ \\

    \midrule
    \multirow{2}{*}{Ouro-2.6B}
        & Original
        & 52.20 & 66.13 & 81.80
        & 79.54 & 76.40 & 71.21 & --
        & $1.00\times$ & $1.00\times$ \\
        & FlashLoop
        & 52.80 & 65.10 & 82.56
        & 79.58 & 75.37 & 71.08 & $-0.13$
        & $1.64\times$ & $6.06\times$ \\

    \midrule
    \multirow{2}{*}{Ouro-2.6B-Thinking}
        & Original
        & 59.00 & 66.21 & 83.47
        & 78.22 & 76.09 & 72.60 & --
        & $1.00\times$ & $1.00\times$ \\
        & FlashLoop
        & 55.60 & 67.38 & 83.94
        & 79.25 & 75.03 & 72.24 & $-0.36$
        & $1.64\times$ & $6.06\times$ \\

    \midrule
    \multirow{2}{*}{Huginn-3.5B}
        & Original
        & 11.00 & 43.94 & 27.52
        & 68.01 & 62.51 & 42.60 & --
        & $1.00\times$ & $1.00\times$ \\
        & FlashLoop
        & 13.40 & 44.45 & 25.70
        & 67.95 & 62.35 & 42.77 & $+0.17$
        & $1.52\times$ & $5.18\times$ \\

    \bottomrule
\end{tabular}%
}
\end{table*}
\subsection{Ablation Study}
\subsubsection{Contribution of Each Component}
We conduct a series of ablation experiments to assess the contribution of each component in FlashLoop. Our goal is to maximize efficiency while preserving the accuracy of looped Transformers, so we evaluate each component's impact on accuracy and its contribution to efficiency.

\paragraph{Accuracy.}
Table~\ref{tab:flashloop_component_ablation} reports the accuracy ablation results. Loop-aware sparse attention alone slightly improves the average performance over the original model. Adding token-sparse updates largely preserves this gain and remains competitive across all three benchmarks, suggesting that the introduced sparsity does not compromise model accuracy. KV residual quantization introduces a small accuracy drop, while the full FlashLoop configuration still maintains performance comparable to the original model. We further ablate the design of KV quantization by replacing cross-loop residual quantization with direct quantization of the full KV states under the same quantization setting. This alternative leads to a noticeably larger accuracy degradation, consistent with our observation in Section~\ref{sec:residual_friendly} that cross-loop KV residuals are more quantization-friendly than full KV states.

\begin{table*}[t]
\centering

\begin{minipage}[t]{0.54\textwidth}
\vspace{0pt}
\centering
\caption{\textbf{Component ablation of FlashLoop on Ouro-1.4B.}
FlashLoop w/ Per-loop KIVI4 means FlashLoop with full KV cache quantization along the same quantization axis. All values are accuracy (\%).}
\label{tab:flashloop_component_ablation}

\small
\setlength{\tabcolsep}{3.2pt}
\resizebox{\linewidth}{!}{%
\begin{tabular}{@{}lccccc@{}}
    \toprule
    Method & MATH-500 & ARC-C & GSM8K & Avg.  \\
    \midrule
    Original
        & 65.80 & 59.98 & 78.77 & 68.18  \\
    Loop-aware Sparse Attention
        & 69.00 & 60.24 & 77.48 & 68.91  \\
    + Cross-Loop Token-Sparse Updates
        & 66.60 & 60.07 & 79.23 & 68.63  \\
    FlashLoop w/ Per-loop KIVI4
        & 64.40 & 59.87 & 77.10 & 67.12  \\
    \midrule
    FlashLoop
        & 65.40 & 60.67 & 79.61 & 68.56  \\
    \bottomrule
\end{tabular}%
}
\end{minipage}
\hfill
\begin{minipage}[t]{0.44\textwidth}
\vspace{0pt}
\centering
\caption{\textbf{Comparison with alternative KV-cache strategies.}
We report average accuracy (\%) across five benchmarks. "T" represents thinking version. }
\label{tab:alternative-kv-strategies}

\small
\setlength{\tabcolsep}{3.2pt}
\resizebox{\linewidth}{!}{%
\begin{tabular}{@{}lcccc@{}}
    \toprule
    \multirow{2}{*}{Model}
    & \multirow{2}{*}{Original}
    & H$_2$O
    & Last-step
    & \multirow{2}{*}{FlashLoop} \\
    & & ($-75\%$ KV) & ($-75\%$ KV) & \\
    \midrule
    Ouro-1.4B
        & 70.11 & 62.81 & 68.33 & \textbf{70.48} \\
    Ouro-1.4B-T
        & 67.11 & 62.68 & 65.09 & \textbf{66.23} \\
    Ouro-2.6B
        & 71.21 & 67.35 & 69.96 & \textbf{71.08} \\
    Ouro-2.6B-T
        & 72.60 & 68.01 & 70.78 & \textbf{72.24} \\
    Huginn-3.5B
        & 42.60 & 40.68 & 32.18 & \textbf{42.77} \\
    \bottomrule
\end{tabular}%
}
\end{minipage}

\end{table*}

\begin{figure}
    \centering
    \includegraphics[width=1\linewidth]{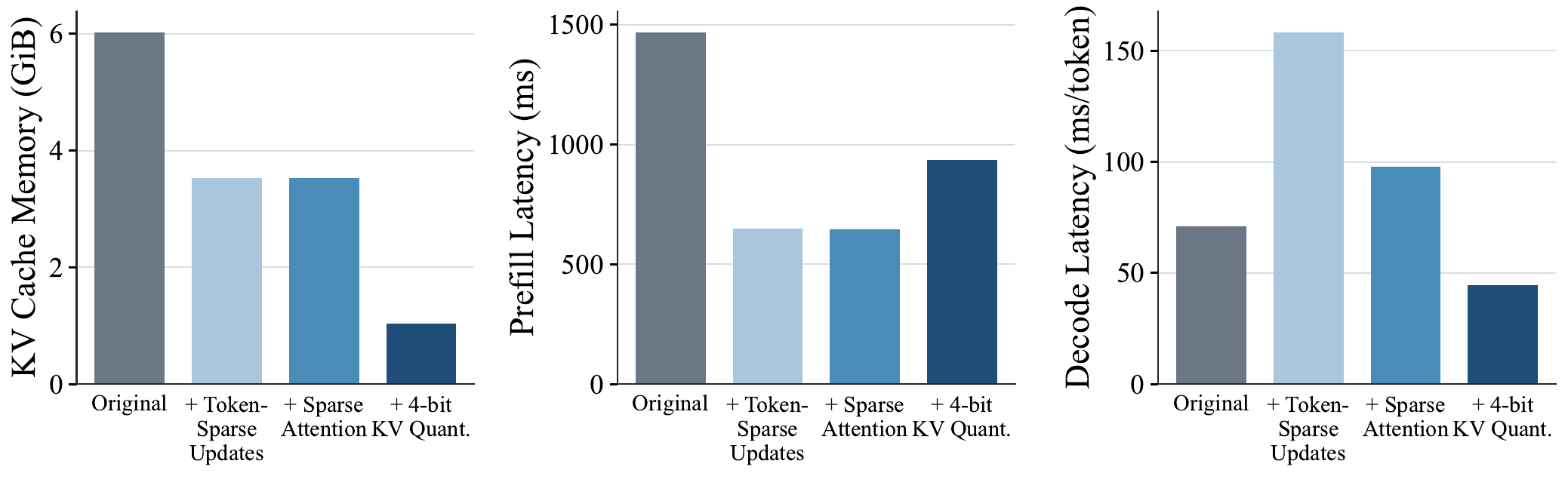}
    \caption{\textbf{Efficiency analysis of each component.} We cumulatively enable cross-loop token-sparse updates, loop-aware sparse attention, and 4-bit KV residual quantization, reporting KV-cache memory, prefill latency, and decode latency. }
    \label{fig:efficiency_ablation}
\end{figure}

\paragraph{Efficiency.}
\looseness=-1 Figure~\ref{fig:efficiency_ablation} reports the efficiency ablation results and shows the distinct benefits of the three components. Cross-loop token-sparse updates substantially reduce prefill latency and KV-cache storage by recomputing only token states that continue to change. Loop-aware sparse attention then offsets much of the resulting decode-latency increase by restricting computation to important key columns. Finally, 4-bit KV residual quantization further reduces cache footprint and decode latency by lowering KV traffic. Although quantization adds work during prefill, the complete FlashLoop configuration is faster than the original model in both prefill and decoding while using substantially less KV memory.

\subsubsection{Comparison with Alternative KV-Cache Strategies}
\label{sec:alternative-kv-strategies}

We further compare FlashLoop with two alternative KV-cache compression strategies. H$_2$O~\citep{zhang2023h2o} retains 25\% of cached tokens based on their accumulated attention scores, while Last-step KV Reuse~\citep{zhu2025scaling} retains only the final-loop KV cache for subsequent decoding, while for Huginn-3.5B we reuse last-step KV of every four steps.

Table~\ref{tab:alternative-kv-strategies} reports the average accuracy across the five benchmarks in Table \ref{tab:main-quality}. H$_2$O compresses each loop independently and therefore does not exploit the structural redundancy across loops, leading to substantial performance degradation. Last-step KV Reuse leverages the recurrent structure but discards the distinct information preserved in intermediate-loop KV states, resulting in additional accuracy loss. FlashLoop instead preserves useful cross-loop information while selectively compressing redundant updates, allowing it to maintain performance close to the original models under substantial KV-cache compression.

\subsubsection{Sparsity and Quantization Configurations}
We conduct parameter ablations to evaluate the sensitivity of FlashLoop to the sparsity and quantization settings of its three components and summarize the resulting accuracy--efficiency trade-offs.

\begin{figure}
    \centering
    \includegraphics[width=1\linewidth]{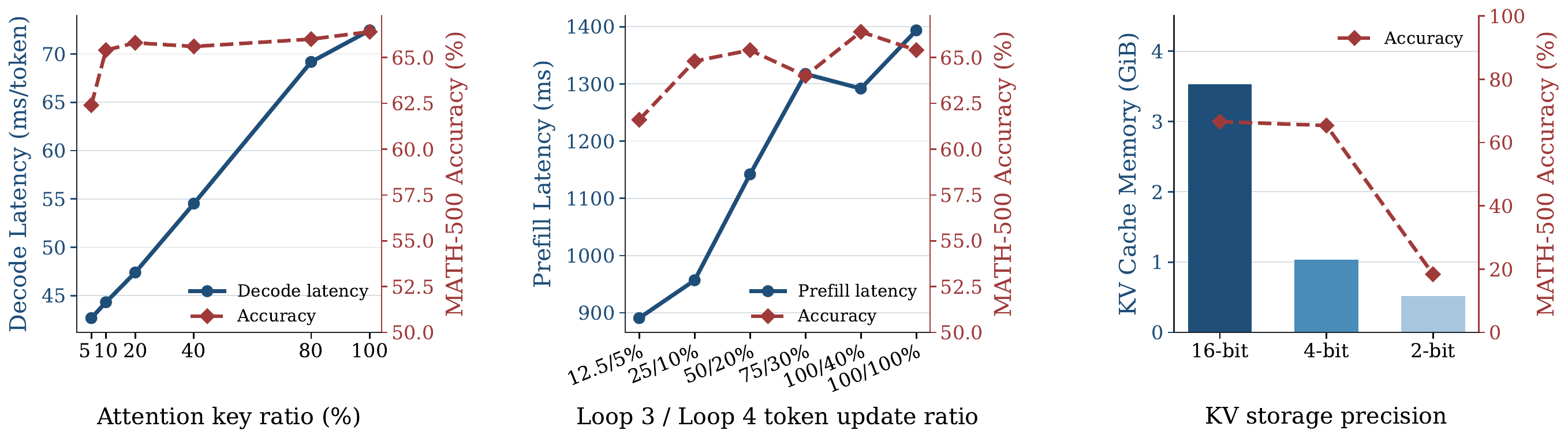}
    \caption{\textbf{Accuracy--efficiency trade-offs of FlashLoop hyperparameters on Ouro-1.4B.} We vary the retained attention-key ratio (left), Loop-3/4 token-update ratios (middle), and KV residual quantization precision (right), while keeping the remaining components fixed. }
    \label{fig:hyperparameter_efficiency_accuracy_8k}
\end{figure}

As reported in Figure~\ref{fig:hyperparameter_efficiency_accuracy_8k}, model accuracy is robust to attention-column sparsity across most evaluated configurations, with noticeable degradation only at very aggressive column sparsity. It remains nearly unchanged from the accuracy of the original model except at 5\% column retention. The model is similarly robust to our cross-loop token-sparse updates: retaining 25\% of tokens in loop 3 and 10\% in loop 4 can maintain accuracy comparable to that of the original model, while keeping system overhead very low.
Moreover, compared with BF16, 4-bit KV residual quantization substantially reduces the KV cache footprint and decode kernel latency, with negligible impact on accuracy, whereas 2-bit setting suffers from clear performance degradation.

\subsection{End-to-end Efficiency Analysis}
We compare baseline inference with the full FlashLoop across context lengths and loops on a single NVIDIA A100-SXM4-40GB GPU.
To improve measurement stability, we repeat prefill three times after warm-up and report the median time measured using CUDA events. For decode, we use 16 tokens for warm-up and measure the average per-token latency over the next 16 tokens. 

\begin{figure}
    \centering
    \includegraphics[width=0.87\linewidth]{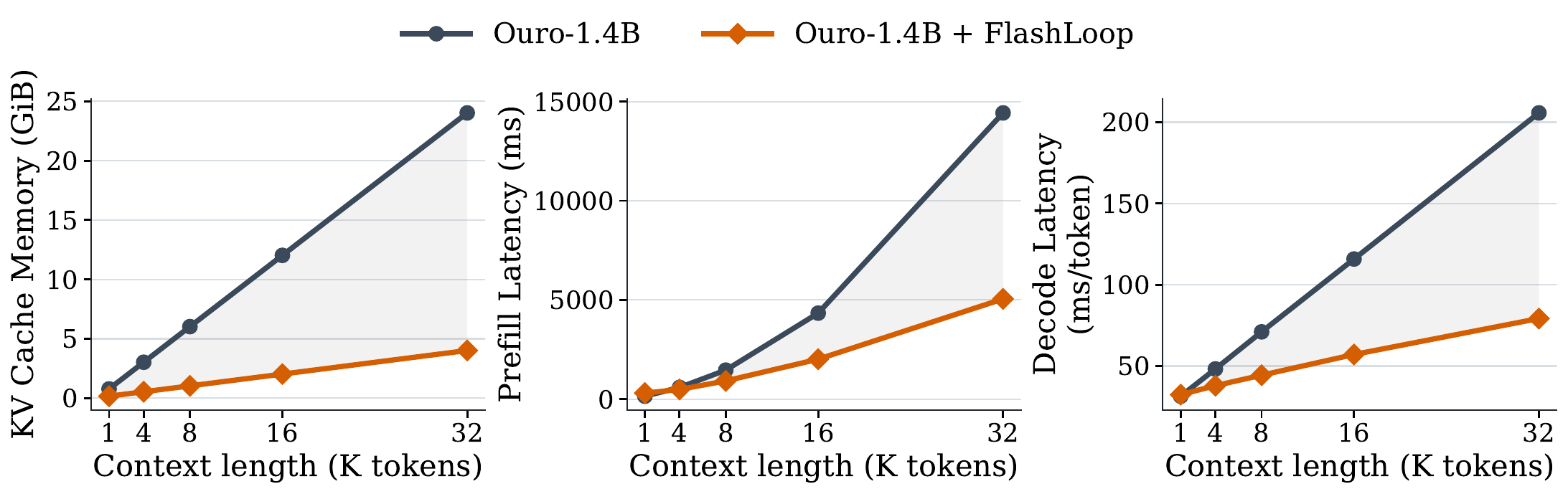}
    \caption{\textbf{Scaling with context length.} FlashLoop delivers increasingly substantial memory savings and speedups as the context length grows.}
    \label{fig:efficiency_context_length}
\end{figure}

\paragraph{Scaling with Context Length.}
As shown in Figure~\ref{fig:efficiency_context_length}, the benefits of FlashLoop become more pronounced as the context length increases. At 32K context length, FlashLoop reduces KV-cache storage by 20 GiB, corresponding to an 82.8\% reduction, while achieving a significant speedup in prefill and decode. We also provide a long-context quality evaluation in Appendix \ref{app:long_context_quality}.

\paragraph{Scaling with Loops.}

\begin{figure}
    \centering
    \includegraphics[width=1\linewidth]{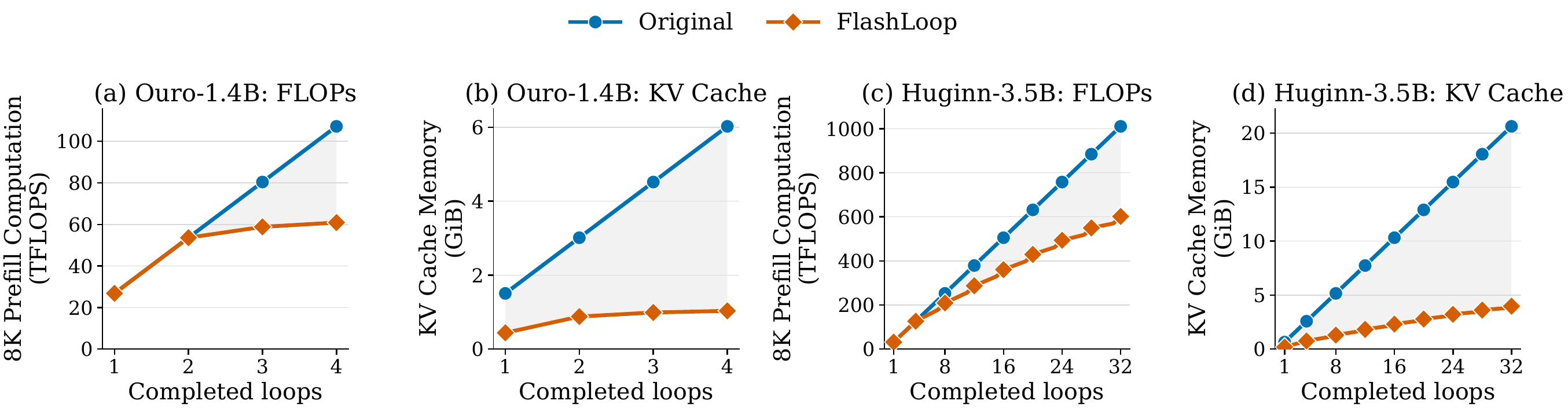}
    \caption{\textbf{Scaling with loops.} As the number of loops increases, FlashLoop progressively reduces the accumulated prefill FLOPs and KV-cache memory of both Ouro-1.4B and Huginn-3.5B.}
    \label{fig:loop_efficiency}
\end{figure}

Figure~\ref{fig:loop_efficiency} shows how computation and KV-cache memory scale with the number of loops for Ouro-1.4B and Huginn-3.5B at an 8K context length. For the original models, both quantities grow approximately linearly with loop count. In contrast, FlashLoop substantially slows this growth in later loops, where only a subset of tokens and attention columns are updated and less additional KV cache is stored. As a result, the efficiency gains become increasingly pronounced as more loops are executed, particularly for Huginn-3.5B with 32 loops. These results demonstrate that FlashLoop effectively mitigates the inference overhead associated with deep recurrence.

\section{Conclusion}

\looseness=-1 In this work, we identify three forms of cross-loop redundancy in looped Transformers, and based on these observations, we propose FlashLoop, a training-free inference framework that exploits such redundancy through cross-loop token-sparse updates, loop-aware sparse attention, and cross-loop KV residual quantization. Across multiple looped models and benchmarks, FlashLoop substantially reduces computation and KV-cache storage while largely preserving the task performance of the original models, while these gains become more pronounced with longer contexts and deeper recurrence.

\bibliography{iclr2027_conference}
\bibliographystyle{iclr2027_conference}
\newpage

\appendix
\section{Performance-matched Looped and Non-looped Models}
\label{app:performance_open_sourced}
We report the accuracy of performance-matched looped and non-looped large language models, including Ouro-2.6B \citep{zhu2025scaling} with four loops, Qwen3-8B \citep{yang2025qwen3}, Gemma3-12B \citep{team2025gemma} and Llama-3.1-8B \citep{grattafiori2024llama}. The results in Table \ref{tab:ouro-paper-comparison} and Figure \ref{fig:pareto_frontier} illustrate that Ouro-2.6B with four loops achieves comparable accuracy in several benchmarks, but its inference efficiency is still a challenge because of its KV cache memory overhead and computation. FlashLoop effectively resolves this challenge without training.
\begin{table*}[t]
\centering
\small
\setlength{\tabcolsep}{7pt}
\caption{\textbf{Benchmark results of open-sourced performance-matched looped and non-looped models.} All models are base models.
The best score in each column is \textbf{bolded}, and the second-best is \underline{underlined}.}
\label{tab:ouro-paper-comparison}
\begin{tabular}{@{}lcccccc@{}}
\toprule
Model & ARC-C  & HellaSwag & WinoGrande & GSM8K & MATH-500 \\
\midrule
Ouro-2.6B R4
& \underline{66.13}  & 79.54 & 76.40
& \underline{81.80} & 52.20 \\
Qwen3-8B
& 66.10  & 79.60 & 76.80
& \textbf{83.09} & \underline{62.30} \\
Gemma3-12B
& \textbf{72.44}  & \textbf{83.68} & \textbf{77.74}
& 77.18 & \textbf{63.20} \\
Llama-3.1-8B
& 60.75  & \underline{81.97} & \underline{77.11}
& 78.17 & 52.90 \\
\bottomrule
\end{tabular}
\end{table*}

\section{Cross-loop redundancy in model with more loops}
\label{app:redundancy_in_more_loops}
\begin{figure}
    \centering
    \includegraphics[width=1\linewidth]{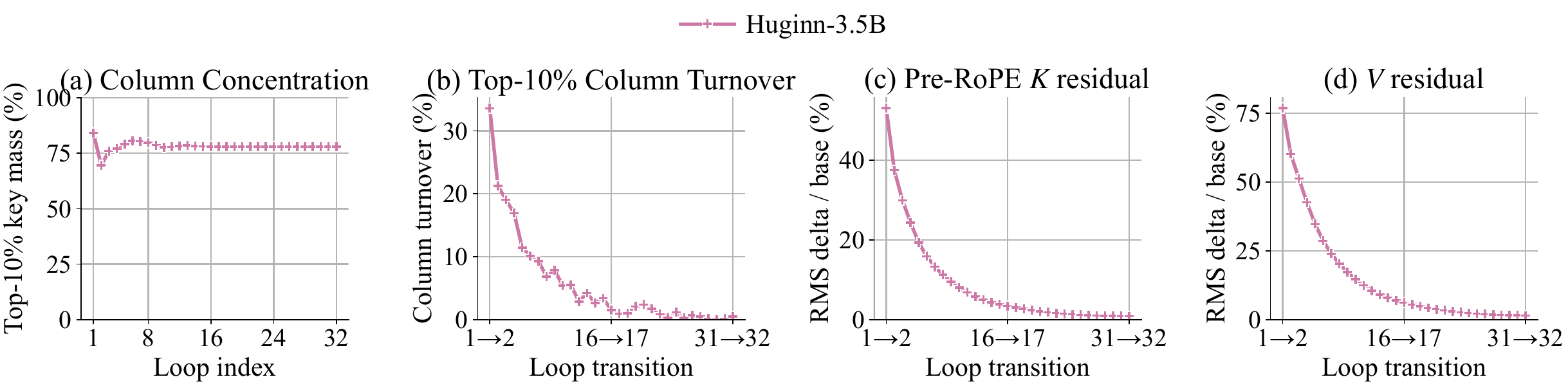}
    \caption{\textbf{Cross-loop redundancy in Huginn-3.5B.} We record the same metrics of Huginn-3.5B as in Figure \ref{fig:cross_loop_redundancy}, which show similar redundancy as in Ouro family.}
    \label{fig:huginn_recurrent_generalization}
\end{figure}

We also study Huginn-3.5B with 32 recurrent loops and observe similar cross-loop redundancy and attention concentration. As shown in Figure~\ref{fig:huginn_recurrent_generalization}, a small fraction of columns captures most of the attention probability mass. In later loops, these important columns become highly stable, while the KV residuals between adjacent loops shrink substantially. These observations show that the structure exploited by FlashLoop also appears in models with much deeper recurrent computation.

\section{Attention Patterns Across Loops}

\begin{figure}
    \centering
    \includegraphics[width=0.9\linewidth]{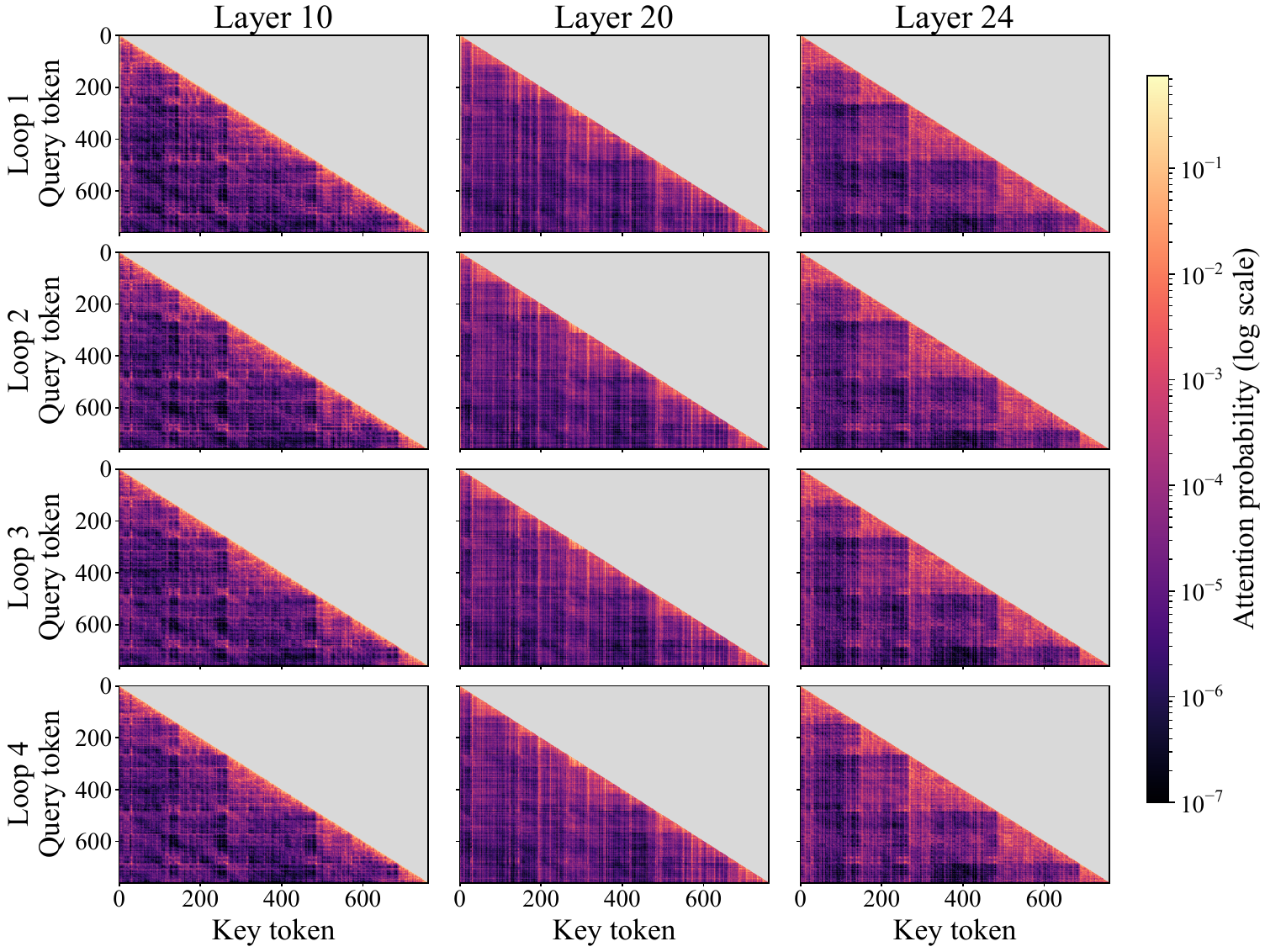}
    \caption{\textbf{Prefill attention maps for a random sampled MATH-500 prompt.} There are obvious important key-column patterns across different layers and loops, while the important columns are stable across loops.}
    \label{fig:appendix_attention_maps}
\end{figure}

Figure~\ref{fig:appendix_attention_maps} visualizes prefill attention probabilities across four loops of Ouro-1.4B. Within the same layer, important key-column patterns recur across loops, and the columns stay stable across loops. This cross-loop persistence motivates FlashLoop to use the preceding loop's attention probabilities to select the columns updated in the next loop.

\section{Mathematical Formulation of FlashLoop}
\label{app:flashloop_math}
We provide the mathematical formulation of our method.
In our formulation,  $H$, $K$, and $V$ denote hidden states, keys, and values. We use
$r$ to index loops and $i,j$ to index token positions. 

\subsection{Cross-Loop Token-Sparse Updates}
\label{app:math_token_updates}

After the dense early loops, FlashLoop selects active tokens using their
normalized hidden-state changes in the preceding loop:
\begin{equation}
\begin{aligned}
s_i
&=\frac{\|H_i^{(r-1)}-H_i^{(r-2)}\|_2}
        {\|H_i^{(r-1)}\|_2+\epsilon},\\
A^{(r)}
&=\operatorname{TopK}_{k^{(r)}}
  \bigl(\{s_i:i\in A^{(r-1)}\}\bigr),
\end{aligned}
\label{eq:app_token_selection}
\end{equation}
where $A^{(r)}$ is the active-token set, $k^{(r)}$ is its budget, and
$\epsilon>0$ is a small constant. After dense early loops update all tokens,
later loops recompute only $A^{(r)}\subseteq A^{(r-1)}$, while the remaining
tokens reuse their hidden states and KV cache. 
\subsection{Loop-aware Sparse Attention}
\label{app:math_sparse_attention}

For the current decode position, $d$ denotes the dimension of each attention
head, $\mathcal C$ denotes the keys computed in loop $r-1$, and $m_{\mathcal C}$ denotes
their cached global mass.
During loop $r-1$, FlashLoop computes attention probabilities
$\widehat p^{(r-1)}$ and uses the known next-loop budget $c^{(r)}$
to select and cache the selected index subset $S$ and its global mass $m$:
\begin{equation}
\begin{aligned}
\widehat p_j^{(r-1)}
&=m_{\mathcal C}
\frac{\exp\!\left(q^{(r-1)}\bigl(\widehat K_j^{(r-1)}\bigr)^\top/\sqrt d\right)}
{\sum_{i\in\mathcal C}
 \exp\!\left(q^{(r-1)}\bigl(\widehat K_i^{(r-1)}\bigr)^\top/\sqrt d\right)},
\quad j\in\mathcal C,\\
S&=\operatorname{TopK}_{c^{(r)}}
  \bigl(\{\widehat p_j^{(r-1)}:j\in\mathcal C\}\bigr),
\qquad
m=\sum_{j\in S}\widehat p_j^{(r-1)},
\end{aligned}
\label{eq:app_column_selection}
\end{equation}
where a dense source loop uses all causal keys and $m_{\mathcal C}=1$.
A sparse source loop uses its selected keys and the mass cached for that
selection, where the budget satisfies $1\leq c^{(r)}\leq|\mathcal C|$.
In loop $r$, FlashLoop reuses the cached subset $S$ and mass $m$ to compute
\begin{equation}
\begin{aligned}
\widehat p_S^{(r)}
&=m\,\operatorname{softmax}\!\left(
  \frac{q^{(r)}\bigl(\widehat K_S^{(r)}\bigr)^\top}{\sqrt d}
  \right),\\
\widehat o^{(r)}
&=\widehat o^{(r-1)}
 -\widehat p_S^{(r-1)}\widehat V_S^{(r-1)}
 +\widehat p_S^{(r)}\widehat V_S^{(r)},
\end{aligned}
\label{eq:app_attention_replacement}
\end{equation}
where $q^{(r)}$ is the current query. Within loop $r$, the updated probabilities
likewise prepare the nested selection and cached mass for loop $r+1$. The reconstruction results in Figure \ref{fig:flashloop_motivation}(b) use this mass-corrected formulation.

\subsection{Cross-Loop KV Residual Quantization}
\label{app:math_kv_quantization}

As described in Section \ref{sec:kv-quantization}, for $X\in\{K,V\}$, FlashLoop stores a quantized base and cross-loop
residuals. 
Let $Q$ and $D$ denote quantization and dequantization. The first loop provides
$\widehat X^{(1)}=D(Q(X^{(1)}))$. For later loops, we calculate $\widehat X_i^{(r)}$, the reconstructed KV used by sparse attention, as follows:
\begin{equation}
\begin{aligned}
\Delta X_i^{(r)}
 &=X_i^{(r)}-\widehat X_i^{(r-1)},\qquad i\in A^{(r)},\\
\widehat X_i^{(r)}
 &=\begin{cases}
 \widehat X_i^{(r-1)}+D\!\left(Q\!\left(\Delta X^{(r)}\right)\right)_i,
       &i\in A^{(r)},\\
 \widehat X_i^{(r-1)},&i\notin A^{(r)}.
 \end{cases}
\end{aligned}
\label{eq:app_masked_residual_codec}
\end{equation}
Only active-token residuals are quantized, and inactive tokens require no new
payload. Overall, token sparsity determines which tokens to update, quantization
compresses them, and column sparsity determines which keys and values are read.

\section{Detailed Experimental Settings}
\label{app:experimental_setting}

\paragraph{Sparsity Setting}
In our experiments, we use the same configuration of FlashLoop for each model on all
benchmarks.    Table~\ref{tab:flashloop_loop_schedules}
summarizes the complete loop-wise sparsity setting used in our experiments.

\paragraph{Quantization Setting}
For both the KV base states and residuals, we use the same quantization scheme. Specifically, we apply INT4 asymmetric group-wise quantization, where the scaling factor and offset are computed from the minimum and maximum values within each group. The group size is set to 64.
Following prior work~\citep{liu2024kivi}, we quantize post-RoPE keys using per-channel quantization, while values are quantized per token. In addition, we keep the most recent 64 tokens in BF16.
In sparse loops, the K/V states generated by active tokens are packed as cross-loop residuals and then quantized using the same scheme described above.

\paragraph{Speedup and KV Cache Memory}
For speedup and KV cache memory measurement, we record the end-to-end wall clock time and peak KV cache memory for the whole benchmark experiments.

\paragraph{Hyperparameter}
We randomly sample 128 sequences from WikiText-2 as a fixed calibration set. For each model, we select the smallest retention ratio such that the relative reconstruction error of the corresponding hidden states or attention outputs remains below 5\%. The resulting loop-wise sparsity schedule is calibrated once per model and then fixed across all downstream tasks and benchmarks.

\begin{table*}
\centering
\caption{\textbf{Loop-wise sparsity configurations used in the main
experiments.} Token retention is the fraction of prompt-token rows
recomputed during prefill. Column retention is the fraction of attention
columns recomputed during decode. }
\label{tab:flashloop_loop_schedules}
\small
\setlength{\tabcolsep}{5pt}
\begin{tabular}{llccl}
\toprule
Model family & Loop(s) & Tokens & Columns & Mode \\
\midrule

\multirow{3}{*}{Ouro-1.4B}
& 1--2 & 100\% & 100\% & Dense warm-up \\
& 3    & 25\%  & 10\%  & Sparse \\
& 4    & 10\%  & 10\%  & Sparse \\

\midrule
\multirow{3}{*}{Ouro-1.4B-Thinking}
& 1--2 & 100\% & 100\% & Dense warm-up \\
& 3    & 25\%  & 12\%  & Sparse \\
& 4    & 10\%  & 12\%  & Sparse \\

\midrule
\multirow{3}{*}{Ouro-2.6B}
& 1--2 & 100\% & 100\% & Dense warm-up \\
& 3    & 20\%  & 8\%   & Sparse \\
& 4    & 8\%   & 8\%   & Sparse \\

\midrule
\multirow{3}{*}{Ouro-2.6B-Thinking}
& 1--2 & 100\% & 100\% & Dense warm-up \\
& 3    & 20\%  & 8\%   & Sparse \\
& 4    & 8\%   & 8\%   & Sparse \\

\midrule
\multirow{2}{*}{Huginn-3.5B}
& 1--4  & 100\% & 100\% & Dense warm-up \\
& 5--32 & $90\%\!\rightarrow\!20\%$
& $76\%\!\rightarrow\!10\%$
& Sparse with smooth linear decay \\
\bottomrule
\end{tabular}
\end{table*}

\section{Long-Context Quality Evaluation}
\label{app:long_context_quality}
To evaluate the effect of FlashLoop in long-context setting, we compare the perplexity of Ouro-1.4B with and without FlashLoop in WikiText-2 \citep{merity2016pointer} in context length of 8K, 16K and 32K tokens. Specifically, we construct a deterministic prefix from the same token stream and evaluate perplexity over the subsequent 128 ground-truth tokens.  Moreover, we also use Needle-in-a-Haystack benchmark to evaluate the long context ability. For FlashLoop we use the same hyperparameters as main experiment. As illustrated in Table \ref{tab:wikitext-ppl}, across different long-context setting, FlashLoop does not bring significant performance degradation.

\begin{table*}[t]
\centering
\small
\setlength{\tabcolsep}{7pt}
\caption{\textbf{Long-context quality evaluation.}
WikiText-2 perplexity is measured on 128 continuation tokens
(lower is better). Needle-in-a-Haystack accuracy is measured over full single-needle and multi-needles samples.}
\label{tab:wikitext-ppl}
\begin{tabular}{@{}lccccc@{}}
\toprule
\multirow{2}{*}{Setting}
& \multicolumn{3}{c}{WikiText-2 PPL $\downarrow$}
& \multicolumn{2}{c}{Needle-in-a-Haystack (\%) $\uparrow$} \\
\cmidrule(lr){2-4}\cmidrule(l){5-6}
& 8K & 16K & 32K
& Single-needle & Multi-needles \\
\midrule
Ouro-1.4B
& 10.526 & \textbf{4.231} & \textbf{5.321} & 83.0 & \textbf{9.2} \\
Ouro-1.4B + FlashLoop
& \textbf{10.513} & 4.266 & 5.589 & \textbf{84.5} & 7.5\\
\bottomrule
\end{tabular}
\end{table*}


\section{Efficient GPU Implementation}
\label{app:open_source}

We implement both a PyTorch reference implementation and an
optimized GPU inference engine. We design the latter to translate
FlashLoop's theoretical computation and memory reductions into actual runtime
and memory savings, avoiding the kernel-launch, memory-traffic, and
intermediate-tensor overheads of a direct PyTorch implementation.

\begin{itemize}[
leftmargin=1.2em,
labelsep=0.4em,
itemsep=1pt,
topsep=2pt,
parsep=0pt,
partopsep=0pt
]
\item \textbf{Sparse execution over active tokens.}
In later loops, we process only selected token rows through Transformer
blocks, while inactive rows directly reuse hidden states and KV entries
from the preceding loop.

\item \textbf{Fused projections.}
We compute Q, K, and V projections with a single GEMM and similarly fuse
the gated MLP's gate and up projections, reducing kernel-launch overhead.

\item \textbf{Physically packed KV cache.}
We store KV states directly in a packed 4-bit representation on the GPU, together with lightweight quantization metadata and only a short full-precision residual tail.

\item \textbf{Attention over compressed cross-loop KV.}
For attention over quantized cross-loop KV, we use custom CUDA kernels
to read packed anchor and residual KV streams directly during QK and PV computation, avoiding explicit reconstruction of a full BF16 KV cache.

\item \textbf{Sparse decode kernels.}
During late-loop decoding, FlashLoop executes attention only on selected key
columns and corresponding value rows. We implement a custom kernel to fuse selection and computation, so that intermediate computation scales
with the retained budget.

\end{itemize}

The engine is specialized for inference with BF16 activations on NVIDIA GPUs. It is intended as a reference implementation of FlashLoop's sparse execution and KV compression techniques. For efficiency evaluation of baseline, we also use inference engine but close KV cache quantization and sparsity configuration, which helps us to fairly evaluate the gain of our method. Both implementations are available at \href{https://github.com/Superone77/FlashLoop}{\texttt{github.com/Superone77/FlashLoop}}.

\end{document}

%% file: math_commands.tex
\usepackage{amsmath,amsfonts,bm}

\def\eqref#1{equation~\ref{#1}}

\def\1{\bm{1}}

\DeclareMathAlphabet{\mathsfit}{\encodingdefault}{\sfdefault}{m}{sl}
\SetMathAlphabet{\mathsfit}{bold}{\encodingdefault}{\sfdefault}{bx}{n}













%% file: iclr2027_conference.bbl
\begin{thebibliography}{28}
\providecommand{\natexlab}[1]{#1}
\providecommand{\url}[1]{\texttt{#1}}
\expandafter\ifx\csname urlstyle\endcsname\relax
  \providecommand{\doi}[1]{doi: #1}\else
  \providecommand{\doi}{doi: \begingroup \urlstyle{rm}\Url}\fi

\bibitem[Brandon et~al.(2024)Brandon, Mishra, Nrusimha, Panda, and
  Kelly]{brandon2024reducing}
William Brandon, Mayank Mishra, Aniruddha Nrusimha, Rameswar Panda, and
  Jonathan~Ragan Kelly.
\newblock Reducing transformer key-value cache size with cross-layer attention,
  2024.
\newblock \emph{URL https://arxiv. org/abs/2405.12981}, 2024.

\bibitem[Cai et~al.(2024)Cai, Zhang, Gao, Liu, Liu, Lu, Xiong, Dong, Chang, Hu,
  et~al.]{cai2024pyramidkv}
Zefan Cai, Yichi Zhang, Bofei Gao, Yuliang Liu, Tianyu Liu, Keming Lu, Wayne
  Xiong, Yue Dong, Baobao Chang, Junjie Hu, et~al.
\newblock Pyramidkv: Dynamic kv cache compression based on pyramidal
  information funneling, 2024.
\newblock \emph{URL https://arxiv. org/abs/2406.02069}, 2024.

\bibitem[Clark et~al.(2018)Clark, Cowhey, Etzioni, Khot, Sabharwal, Schoenick,
  and Tafjord]{allenai_arc}
Peter Clark, Isaac Cowhey, Oren Etzioni, Tushar Khot, Ashish Sabharwal, Carissa
  Schoenick, and Oyvind Tafjord.
\newblock Think you have solved question answering? try arc, the ai2 reasoning
  challenge.
\newblock \emph{arXiv:1803.05457v1}, 2018.

\bibitem[Cobbe et~al.(2021)Cobbe, Kosaraju, Bavarian, Chen, Jun, Kaiser,
  Plappert, Tworek, Hilton, Nakano, Hesse, and Schulman]{cobbe2021gsm8k}
Karl Cobbe, Vineet Kosaraju, Mohammad Bavarian, Mark Chen, Heewoo Jun, Lukasz
  Kaiser, Matthias Plappert, Jerry Tworek, Jacob Hilton, Reiichiro Nakano,
  Christopher Hesse, and John Schulman.
\newblock Training verifiers to solve math word problems.
\newblock \emph{arXiv preprint arXiv:2110.14168}, 2021.

\bibitem[Dehghani et~al.(2018)Dehghani, Gouws, Vinyals, Uszkoreit, and
  Kaiser]{dehghani2018universal}
Mostafa Dehghani, Stephan Gouws, Oriol Vinyals, Jakob Uszkoreit, and {\L}ukasz
  Kaiser.
\newblock Universal transformers.
\newblock \emph{arXiv preprint arXiv:1807.03819}, 2018.

\bibitem[Gao et~al.(2024)Gao, Tow, Abbasi, Biderman, Black, DiPofi, Foster,
  Golding, Hsu, Le~Noac'h, Li, McDonell, Muennighoff, Ociepa, Phang, Reynolds,
  Schoelkopf, Skowron, Sutawika, Tang, Thite, Wang, Wang, and
  Zou]{eval-harness}
Leo Gao, Jonathan Tow, Baber Abbasi, Stella Biderman, Sid Black, Anthony
  DiPofi, Charles Foster, Laurence Golding, Jeffrey Hsu, Alain Le~Noac'h,
  Haonan Li, Kyle McDonell, Niklas Muennighoff, Chris Ociepa, Jason Phang,
  Laria Reynolds, Hailey Schoelkopf, Aviya Skowron, Lintang Sutawika, Eric
  Tang, Anish Thite, Ben Wang, Kevin Wang, and Andy Zou.
\newblock The language model evaluation harness, 07 2024.
\newblock URL \url{https://zenodo.org/records/12608602}.

\bibitem[Geiping et~al.(2026)Geiping, McLeish, Jain, Kirchenbauer, Singh,
  Bartoldson, Kailkhura, Bhatele, and Goldstein]{geiping2026scaling}
Jonas Geiping, Sean McLeish, Neel Jain, John Kirchenbauer, Siddharth Singh,
  Brian Bartoldson, Bhavya Kailkhura, Abhinav Bhatele, and Tom Goldstein.
\newblock Scaling up test-time compute with latent reasoning: A recurrent depth
  approach.
\newblock \emph{Advances in Neural Information Processing Systems},
  38:\penalty0 41340--41391, 2026.

\bibitem[Grattafiori et~al.(2024)Grattafiori, Dubey, Jauhri, Pandey, Kadian,
  Al-Dahle, Letman, Mathur, Schelten, Vaughan, et~al.]{grattafiori2024llama}
Aaron Grattafiori, Abhimanyu Dubey, Abhinav Jauhri, Abhinav Pandey, Abhishek
  Kadian, Ahmad Al-Dahle, Aiesha Letman, Akhil Mathur, Alan Schelten, Alex
  Vaughan, et~al.
\newblock The llama 3 herd of models.
\newblock \emph{arXiv preprint arXiv:2407.21783}, 2024.

\bibitem[Hooper et~al.(2024)Hooper, Kim, Mohammadzadeh, Mahoney, Shao, Keutzer,
  and Gholami]{hooper2024kvquant}
Coleman Hooper, Sehoon Kim, Hiva Mohammadzadeh, Michael~W Mahoney, Yakun~S
  Shao, Kurt Keutzer, and Amir Gholami.
\newblock Kvquant: Towards 10 million context length llm inference with kv
  cache quantization.
\newblock \emph{Advances in Neural Information Processing Systems},
  37:\penalty0 1270--1303, 2024.

\bibitem[Jiang et~al.(2024)Jiang, Li, Zhang, Wu, Luo, Ahn, Han, Abdi, Li, Lin,
  et~al.]{jiang2024minference}
Huiqiang Jiang, Yucheng Li, Chengruidong Zhang, Qianhui Wu, Xufang Luo, Surin
  Ahn, Zhenhua Han, Amir~H Abdi, Dongsheng Li, Chin-Yew Lin, et~al.
\newblock Minference 1.0: Accelerating pre-filling for long-context llms via
  dynamic sparse attention.
\newblock \emph{Advances in Neural Information Processing Systems},
  37:\penalty0 52481--52515, 2024.

\bibitem[Li et~al.(2024)Li, Huang, Yang, Venkitesh, Locatelli, Ye, Cai, Lewis,
  and Chen]{li2024snapkv}
Yuhong Li, Yingbing Huang, Bowen Yang, Bharat Venkitesh, Acyr Locatelli,
  Hanchen Ye, Tianle Cai, Patrick Lewis, and Deming Chen.
\newblock Snapkv: Llm knows what you are looking for before generation.
\newblock \emph{Advances in Neural Information Processing Systems},
  37:\penalty0 22947--22970, 2024.

\bibitem[Lightman et~al.(2023)Lightman, Kosaraju, Burda, Edwards, Baker, Lee,
  Leike, Schulman, Sutskever, and Cobbe]{lightman2023lets}
Hunter Lightman, Vineet Kosaraju, Yura Burda, Harri Edwards, Bowen Baker, Teddy
  Lee, Jan Leike, John Schulman, Ilya Sutskever, and Karl Cobbe.
\newblock Let's verify step by step.
\newblock \emph{arXiv preprint arXiv:2305.20050}, 2023.

\bibitem[Liu et~al.(2024{\natexlab{a}})Liu, Liu, Pan, He, Haffari, and
  Zhuang]{liu2024minicache}
Akide Liu, Jing Liu, Zizheng Pan, Yefei He, Gholamreza Haffari, and Bohan
  Zhuang.
\newblock Minicache: Kv cache compression in depth dimension for large language
  models.
\newblock \emph{Advances in Neural Information Processing Systems},
  37:\penalty0 139997--140031, 2024{\natexlab{a}}.

\bibitem[Liu et~al.(2023)Liu, Desai, Liao, Wang, Xie, Xu, Kyrillidis, and
  Shrivastava]{liu2023scissorhands}
Zichang Liu, Aditya Desai, Fangshuo Liao, Weitao Wang, Victor Xie, Zhaozhuo Xu,
  Anastasios Kyrillidis, and Anshumali Shrivastava.
\newblock Scissorhands: Exploiting the persistence of importance hypothesis for
  llm kv cache compression at test time.
\newblock \emph{Advances in Neural Information Processing Systems},
  36:\penalty0 52342--52364, 2023.

\bibitem[Liu et~al.(2024{\natexlab{b}})Liu, Yuan, Jin, Zhong, Xu, Braverman,
  Chen, and Hu]{liu2024kivi}
Zirui Liu, Jiayi Yuan, Hongye Jin, Shaochen Zhong, Zhaozhuo Xu, Vladimir
  Braverman, Beidi Chen, and Xia Hu.
\newblock Kivi: A tuning-free asymmetric 2bit quantization for kv cache.
\newblock \emph{arXiv preprint arXiv:2402.02750}, 2024{\natexlab{b}}.

\bibitem[Merity et~al.(2016)Merity, Xiong, Bradbury, and
  Socher]{merity2016pointer}
Stephen Merity, Caiming Xiong, James Bradbury, and Richard Socher.
\newblock Pointer sentinel mixture models, 2016.

\bibitem[Neill \& Reid(2026)Neill and Reid]{neill2026looped}
James~O' Neill and Fergal Reid.
\newblock Looped latent attention: Cross-loop kv compression for looped
  transformers.
\newblock \emph{arXiv preprint arXiv:2607.15456}, 2026.

\bibitem[Ribar et~al.(2023)Ribar, Chelombiev, Hudlass-Galley, Blake, Luschi,
  and Orr]{ribar2023sparq}
Luka Ribar, Ivan Chelombiev, Luke Hudlass-Galley, Charlie Blake, Carlo Luschi,
  and Douglas Orr.
\newblock Sparq attention: Bandwidth-efficient llm inference.
\newblock \emph{arXiv preprint arXiv:2312.04985}, 2023.

\bibitem[Sakaguchi et~al.(2019)Sakaguchi, Bras, Bhagavatula, and
  Choi]{sakaguchi2019winogrande}
Keisuke Sakaguchi, Ronan~Le Bras, Chandra Bhagavatula, and Yejin Choi.
\newblock Winogrande: An adversarial winograd schema challenge at scale.
\newblock \emph{arXiv preprint arXiv:1907.10641}, 2019.

\bibitem[Tang et~al.(2024)Tang, Zhao, Zhu, Xiao, Kasikci, and
  Han]{tang2024quest}
Jiaming Tang, Yilong Zhao, Kan Zhu, Guangxuan Xiao, Baris Kasikci, and Song
  Han.
\newblock Quest: Query-aware sparsity for efficient long-context llm inference.
\newblock \emph{arXiv preprint arXiv:2406.10774}, 2024.

\bibitem[Team et~al.(2025)Team, Kamath, Ferret, Pathak, Vieillard, Merhej,
  Perrin, Matejovicova, Ram{\'e}, Rivi{\`e}re, et~al.]{team2025gemma}
Gemma Team, Aishwarya Kamath, Johan Ferret, Shreya Pathak, Nino Vieillard,
  Ramona Merhej, Sarah Perrin, Tatiana Matejovicova, Alexandre Ram{\'e},
  Morgane Rivi{\`e}re, et~al.
\newblock Gemma 3 technical report.
\newblock \emph{arXiv preprint arXiv:2503.19786}, 2025.

\bibitem[Vendrell et~al.(2026)Vendrell, Masdemont, Grillo, Ros-Giralt,
  Behboodi, and Massoli]{vendrell2026memory}
Victor~Conchello Vendrell, Arnau~Padres Masdemont, Niccol{\`o} Grillo, Jordi
  Ros-Giralt, Arash Behboodi, and Fabio~Valerio Massoli.
\newblock Memory-efficient looped transformer: Decoupling compute from memory
  in looped language models.
\newblock \emph{arXiv preprint arXiv:2605.07721}, 2026.

\bibitem[Wang et~al.(2025)Wang, Pan, Yao, Csordas, Li, Yin, Wu, Zhang, Li, and
  Liu]{wang2025chain}
Zihan Wang, Rui Pan, Jiarui Yao, Robert Csordas, Linjie Li, Lu~Yin, Jiajun Wu,
  Tong Zhang, Manling Li, and Shiwei Liu.
\newblock Chain-of-experts: unlocking the communication power of
  mixture-of-experts models.
\newblock \emph{arXiv preprint arXiv:2506.18945}, 2025.

\bibitem[Yang et~al.(2025)Yang, Li, Yang, Zhang, Hui, Zheng, Yu, Gao, Huang,
  Lv, et~al.]{yang2025qwen3}
An~Yang, Anfeng Li, Baosong Yang, Beichen Zhang, Binyuan Hui, Bo~Zheng, Bowen
  Yu, Chang Gao, Chengen Huang, Chenxu Lv, et~al.
\newblock Qwen3 technical report.
\newblock \emph{arXiv preprint arXiv:2505.09388}, 2025.

\bibitem[Yang et~al.(2026)Yang, Ma, Conzelmann, Zheng, Mahoney, Rusch, and
  Liu]{yang2026alphaq}
Wanqi Yang, Yuexiao Ma, Alexander Conzelmann, Xiawu Zheng, Michael~W Mahoney,
  T~Konstantin Rusch, and Shiwei Liu.
\newblock Alphaq: Calibration-free bit allocation for mixture-of-experts
  quantization.
\newblock \emph{arXiv preprint arXiv:2606.04980}, 2026.

\bibitem[Zellers et~al.(2019)Zellers, Holtzman, Bisk, Farhadi, and
  Choi]{zellers2019hellaswag}
Rowan Zellers, Ari Holtzman, Yonatan Bisk, Ali Farhadi, and Yejin Choi.
\newblock Hellaswag: Can a machine really finish your sentence?
\newblock In \emph{Proceedings of the 57th Annual Meeting of the Association
  for Computational Linguistics}, 2019.

\bibitem[Zhang et~al.(2023)Zhang, Sheng, Zhou, Chen, Zheng, Cai, Song, Tian,
  R{\'e}, Barrett, et~al.]{zhang2023h2o}
Zhenyu Zhang, Ying Sheng, Tianyi Zhou, Tianlong Chen, Lianmin Zheng, Ruisi Cai,
  Zhao Song, Yuandong Tian, Christopher R{\'e}, Clark Barrett, et~al.
\newblock H2o: Heavy-hitter oracle for efficient generative inference of large
  language models.
\newblock \emph{Advances in neural information processing systems},
  36:\penalty0 34661--34710, 2023.

\bibitem[Zhu et~al.(2025)Zhu, Wang, Hua, Zhang, Li, Que, Wei, Wen, Yin, Xing,
  et~al.]{zhu2025scaling}
Rui-Jie Zhu, Zixuan Wang, Kai Hua, Tianyu Zhang, Ziniu Li, Haoran Que, Boyi
  Wei, Zixin Wen, Fan Yin, He~Xing, et~al.
\newblock Scaling latent reasoning via looped language models.
\newblock \emph{arXiv preprint arXiv:2510.25741}, 2025.

\end{thebibliography}
